\documentclass[letterpaper, 10pt, conference]{IEEEtran}

\usepackage{mathtools, amssymb, bm}

\usepackage{graphicx}
\usepackage[utf8]{inputenc}
\usepackage{xcolor}
\usepackage{booktabs}
\usepackage{multirow}
\usepackage{makecell}
\usepackage{subcaption}

\usepackage{algorithm}
\usepackage{algpseudocode}

\usepackage{tikz}
\usepackage{pgfplots}
\usepgfplotslibrary{groupplots}
\usetikzlibrary{arrows, shapes, patterns, decorations.pathmorphing, 
                decorations.pathreplacing, decorations.shapes, 
                decorations.text, decorations.markings, arrows.meta, 
                positioning, shapes.misc, fit, backgrounds}

\usepackage{siunitx}
\usepackage{fp}
\usepackage[nolist]{acronym}
\usepackage{verbatim}
\usepackage{textcomp}
\usepackage{eso-pic}
\usepackage[numbers,sort&compress]{natbib}

\usepackage{authblk}

\usepackage[hyphens]{url}
\usepackage{breakurl}

\usepackage{hyperref}

\usepackage[capitalize]{cleveref}

\definecolor{red}{rgb}{0.5,0,0}
\definecolor{green}{rgb}{0.4660,0.6740,0.1880}
\definecolor{blue}{rgb}{0,0.5,0.75}
\definecolor{mycolor1}{rgb}{0.4940,0.1840,0.5560}%
\definecolor{mycolor2}{rgb}{1,0.5,0}%
\definecolor{mycolor3}{rgb}{0 0 0}%

\newcommand\mydots{\hbox to 0.75em{.\hss.\hss.}}

\IEEEoverridecommandlockouts                              

\title{\LARGE \bf
Knowledge Graph Sparsification for GNN-based Rare Disease Diagnosis
}

\author[1]{Premt Cara$^*$}
\author[1,2]{Kamilia Zaripova$^*$\thanks{$^*$ Equal contribution. Corresponding author: \texttt{p.cara@tum.de}}}
\author[1,2]{David Bani-Harouni}
\author[1,2,3]{Nassir Navab}
\author[1,2]{Azade Farshad}

\affil[1]{Department of Computer Science, Technical University of Munich}
\affil[2]{Munich Center for Machine Learning (MCML), Munich, Germany}
\affil[3]{Johns Hopkins University, Baltimore, Maryland, USA}

\def\methodName{RareNet}

\begin{document}

\begin{acronym}
\acro{com}[COM]{center of mass}
\end{acronym}

\maketitle

\thispagestyle{empty}
\pagestyle{empty}

\begin{abstract}

Rare genetic disease diagnosis faces critical challenges: insufficient patient data, inaccessible full genome sequencing, and the immense number of possible causative genes. These limitations cause prolonged diagnostic journeys, inappropriate treatments, and critical delays, disproportionately affecting patients in resource-limited settings where diagnostic tools are scarce. We propose RareNet, a subgraph-based Graph Neural Network that requires only patient phenotypes to identify the most likely causal gene and retrieve focused patient subgraphs for targeted clinical investigation. RareNet can function as a standalone method or serve as a pre-processing or post-processing filter for other candidate gene prioritization methods, consistently enhancing their performance while potentially enabling explainable insights. Through comprehensive evaluation on two biomedical datasets, we demonstrate competitive and robust causal gene prediction and significant performance gains when integrated with other frameworks. By requiring only phenotypic data, which is readily available in any clinical setting, RareNet democratizes access to sophisticated genetic analysis, offering particular value for underserved populations lacking advanced genomic infrastructure.

\end{abstract}

\setlength{\textfloatsep}{10pt}
\setlength{\floatsep}{10pt}


\section{Introduction}
Rare genetic diseases, though individually uncommon, collectively affect over 300 million people worldwide. For many patients, especially those in underserved or resource-limited settings, access to timely and accurate diagnosis remains out of reach \cite{marwaha2022guide}. Diagnostic odysseys often span years, requiring access to full genome sequencing, expert-curated candidate gene lists, and highly specialized knowledge-resources that are unevenly distributed globally. These disparities result in delayed or missed diagnoses, inappropriate interventions, and irreversible health consequences. 
The challenge is compounded by the vast number of potential causative genes and the often incomplete or inconsistently documented clinical phenotype data-particularly in low-resource settings where standardized medical records may be lacking. Recent advances in AI-driven rare disease diagnostics have shown promise but often depend on comprehensive genomic data, expensive sequencing technologies, and large-scale annotated datasets. As a result, they risk reinforcing existing healthcare inequities by primarily serving patients in well-resourced institutions. In this work, we aim to shift this paradigm. Assuming that the underlying biomedical knowledge graph contains sufficient relationships to support causal gene prediction from phenotypic input, we propose RareNet. RareNet is a subgraph-based Graph Neural Network that identifies plausible causal genes using only phenotypic input--data that is typically available in any clinical encounter, regardless of access to genetic testing infrastructure. RareNet not only predicts the most likely causal gene but also retrieves patient-specific subgraphs, highlighting relevant molecular and phenotypic interactions that may inform further clinical investigation.
We hypothesize that RareNet can serve as a standalone diagnostic tool or enhance existing gene prioritization methods by generating phenotype-only candidate gene lists for pre- or post-filtering. It reduces reliance on curated inputs, handles incomplete or imprecise phenotype data, and incorporates broader clinical context. As a pre-filter, it narrows the search space; as a post-filter, it re-ranks results to improve accuracy. From a social impact perspective, RareNet addresses a foundational barrier in rare disease diagnosis: the lack of accessible, scalable tools for gene prioritization in data-scarce contexts. 
By reducing dependence on genome sequencing and expert curation, our method supports more equitable diagnostic workflows-particularly in regions with limited access to molecular diagnostics. This shift toward phenotype-driven analysis enables earlier interventions, more targeted follow-ups, and improved outcomes across socioeconomic and geographic boundaries. Our key contributions can be summarized as follows: (1) We introduce the first knowledge graph–based method that, using only patient phenotypes, generates a patient-specific subgraph and candidate gene list, and performs causal gene prioritization based on this subgraph. (2) We demonstrate that our approach can function both as a standalone diagnostic tool and as an enhancement to existing gene prioritization methods, enabling their application in settings where curated candidate gene lists are unavailable and enhancing their performance. We show that the generated candidate gene lists can serve as effective pre-filtering or post-filtering modules. (3) We develop a GNN-based architecture that performs robustly even in challenging scenarios with noisy or unreliable phenotype data-conditions under which many state-of-the-art methods fail. The source code will be made public upon acceptance.
\section{Related Work}
\label{sec:related_work}
\paragraph{Rare Disease Prediction.}
Early computational approaches for rare disease diagnosis fall into three categories: genotype-based, phenotype-based, and hybrid methods. While effective in well-resourced settings, these approaches often assume access to high-quality sequencing data, curated candidate gene lists, and expert interpretation-resources that are often unavailable in low-resource clinical environments. Genotype-based tools such as MutationTaster \cite{steinhaus2021mutationtaster2021}, CADD \cite{rentzsch2019cadd}, and M-CAP \cite{jagadeesh2016m} assess variant pathogenicity using genome-wide annotations, but rely on access to well-annotated variants and sequencing infrastructure. Phenotype-driven models like Phenolyzer \cite{yang2015phenolyzer}, Phrank \cite{jagadeesh2019phrank}, and CADA \cite{peng2021cada} use Human Phenotype Ontology (HPO) terms to prioritize genes based on phenotypic similarity. PhenoApt~\cite{cui2020conan} constructs a heterogeneous knowledge graph from resources such as HPO \cite{10.1093/nar/gkaa1043}, OMIM \cite{10.1093/nar/gky1151}, and Orphanet \cite{weinreich2008orphanet}, integrating gene–phenotype and phenotype–phenotype relationships, and uses graph embedding techniques to rank candidate genes based on their similarity to the patient phenotype profile, with support for both clinician-defined and automatically derived phenotype weights. These models, however, often depend on curated gene–phenotype associations and perform poorly when input phenotypes are sparse, noisy, or atypical. Hybrid methods integrate both genomic and phenotypic information. Amelie \cite{birgmeier2020amelie} uses natural language processing to mine the biomedical literature and rank genes from a candidate list based on literature support for gene–phenotype associations. Shepherd \cite{shepherd} employs a graph neural network to combine patient phenotypes with a biomedical knowledge graph and rank genes from a curated candidate list, typically based on rare or pathogenic variants identified through sequencing. Other hybrid frameworks include Exomiser \cite{smedley2015next}, Xrare \cite{li2019xrare}, and AI-MARRVEL \cite{mao2024ai}. However, many of these tools rely on predefined candidate gene lists or Variant Call Format (VCF) files, which require sequencing and expert interpretation-resources that are rarely available in under-resourced settings. Additionally, curated candidate lists introduce biases toward well-characterized genes, limiting generalization to novel or atypical cases.

\paragraph{Knowledge Graphs in Healthcare.}
To move beyond fixed candidate sets, biomedical knowledge graphs (KGs) such as DisGeNET \cite{10.1093/nar/gkz1021}, DRKG \cite{drkg2020}, GenomicKB \cite{10.1093/nar/gkac957}, and PRIME-KG \cite{Chandak2022.05.01.489928} have emerged as comprehensive resources for representing gene–disease–phenotype–drug relationships at scale. KG embedding methods and multimodal frameworks \cite{vilela2023biomedical, ektefaie2023multimodal, galkin2023towards} enable systematic reasoning and support downstream tasks such as disease–gene association prediction and drug repurposing. While these approaches offer a scalable alternative to manual curation, applying KGs directly to patient-level diagnosis remains difficult, particularly in low-resource settings, where both computational capacity and detailed clinical annotations are often limited. Additionally, the scale and heterogeneity of biomedical KGs pose challenges for generating interpretable, patient-specific outputs in real-time clinical workflows.

\paragraph{Subgraph Extraction and KG-Augmented Reasoning.}
Recent advances have sought to improve efficiency and interpretability by extracting concise, context- or patient-specific subgraphs from large knowledge graphs. SubGNN \cite{alsentzer2020subgraph} and PullNet \cite{sun2019pullnet} formalize supervised subgraph prediction and iterative question-specific retrieval, while other methods leverage data augmentation \cite{shen2022improving} or simulate undiagnosed patients \cite{alsentzer2023simulation}. In parallel, hybrid LLM–KG pipelines-such as GNN-RAG \cite{mavromatis2024gnnraggraphneuralretrieval}, G-Retriever \cite{NEURIPS2024_efaf1c97}, MindMap \cite{wen2024mindmapknowledgegraphprompting}, RoG \cite{luo2024reasoninggraphsfaithfulinterpretable}, and GoG \cite{li2023graphreasoningquestionanswering}-retrieve KG subgraphs to support LLM-based reasoning. These methods address challenges such as graph-to-text encoding \cite{fatemi2023talklikegraphencoding, 10387715, 10697304} and hallucination control \cite{Hager2024, galkin2023towards}. However, rare disease diagnosis presents fundamentally different constraints: data is sparse, phenotypes are often noisy or underspecified, and labeled examples are limited. In this setting, existing LLM–KG pipelines, designed for broad, well-resourced domains, struggle to generalize and are prone to hallucinations, making them unreliable for accurate and equitable gene prioritization in real-world clinical use.

\paragraph{Research Gap.}
Existing models often rely on curated gene lists or variant-level inputs, limiting their applicability in low-resource environments. RareNet addresses this gap by operating directly on phenotype-only inputs, without requiring sequencing or expert-curated candidates. It is the first method to jointly extract patient-specific subgraphs from a biomedical knowledge graph, generate candidate gene lists, and prioritize causal genes, opening the paths to interpretable rare disease diagnostics in settings with limited infrastructure.

\section{Methodology}
\begin{figure*}[t]
    \centering
    \begin{tikzpicture}[scale=0.2, 
      font=\tiny	,
      node distance=0.2cm,
      line width=0.1pt,
      standard-arrow/.style={
        -{Triangle[angle=45:3pt 2]},
        line width=0.7pt,
        draw=gray!80
      },
      dashed-arrow/.style={
        -{Triangle[angle=45:3pt 2]},
        line width=0.7pt,
        draw=gray!80,
        dashed
      },
      phenotype-color/.style={
        fill=orange!20,
        draw=orange!60!black,
        line width=0.7pt
      },
      neighbor-color/.style={
        fill=blue!15,
        draw=blue!60!black,
        line width=0.7pt
      },
      subgraph-color/.style={
        fill=blue!15,
        draw=blue!60!black,
        line width=0.7pt
      },
      gat-color/.style={
        fill=teal!15,
        draw=teal!60!black,
        line width=0.7pt
      },
      retrieval-color/.style={
        fill=purple!15,
        draw=purple!60!black,
        line width=0.7pt
      },
      patient-repr-color/.style={
        fill=teal!15,
        draw=teal!60!black,
        line width=0.7pt
      },
      edge-score-color/.style={
        fill=teal!15,
        draw=teal!60!black,
        line width=0.7pt
      },
      gene-score-color/.style={
        fill=teal!15,
        draw=teal!60!black,
        line width=0.7pt
      },
      box/.style={
        rectangle,
        rounded corners=2pt,
        fill=white,
        minimum width=2.5cm,
        minimum height=0.9cm,
        align=center
      },
      smallbox/.style={
        rectangle,
        rounded corners=2pt,
        fill=white,
        minimum width=2.5cm,
        minimum height=0.9cm,
        align=center
      },
      highbox/.style={
        rectangle,
        rounded corners=2pt,
        fill=white,
        minimum width=2.0cm,
        minimum height=4.4cm,
        align=center
      },
      dashedbox/.style={
        rectangle,
        draw=teal!60!black,
        line width=0.8pt,
        dash pattern=on 4pt off 2pt,
        rounded corners=2pt,
        fill=teal!3,
        inner sep=4mm
      }
    ]
    \node[box, phenotype-color, xshift=-2cm] (phenos) {
      \scriptsize \textbf{Patient Phenotypes}\\
      \tiny HP1, HP2, \dots
    };
    \node[box, neighbor-color, below=0.3cm of phenos] (loader) {
      \scriptsize \textbf{Phenotype} \\ \scriptsize \textbf{ Graph Sampling}\\
      \tiny Sample Subgraph
    };
    \draw[standard-arrow] (phenos) -- (loader);
    %
    
    \node[smallbox, gat-color, right=1.8cm of loader] (gat) {
      \scriptsize \textbf{GAT Layers}\\
      \scriptsize $h^{(l)}_i$, $\alpha_{ij}$
    };
    %
    \node[smallbox, patient-repr-color, right=0.8cm of gat, yshift=1.9cm] (scaleddot) {
      \scriptsize \textbf{Patient Repr.}\\
      \scriptsize $\mathbf{p}$
    };
    \draw[standard-arrow] (gat.north) |- (scaleddot.west) node[pos=0.4, right, font=\scriptsize] {$h^{(l)}_i$};
    
    \node[smallbox, edge-score-color, right=0.8cm of gat, yshift=-1.9cm] (edgescore) {
      \scriptsize \textbf{Edge Scorer} \\
      \scriptsize $s_{edge}$ 
    };
    \draw[standard-arrow] (gat.south) |- (edgescore.west) node[pos=0.4, right, font=\scriptsize] {$h^{(l)}_i$, $\alpha_{ij}$};
    
    \node[smallbox, gene-score-color, right=3.5cm of gat] (gene) {
      \scriptsize \textbf{Gene Scorer}\\
      \scriptsize $s_{gene}$
    };
    \draw[standard-arrow] (gat.east) -- (gene.west) node[pos=0.2, above, font=\scriptsize] {$h^{(l)}_i$};
    
    \draw[standard-arrow] (loader.east) -- (gat.west) node[pos=0.25, above, font=\scriptsize] {$SG$};
    
    \draw[standard-arrow] (scaleddot.east) -| (gene.north) node[pos=0.12, below, font=\scriptsize] {$\mathbf{p}$};
    \draw[standard-arrow] (scaleddot.south) -- (edgescore.north) node[pos=0.1, left, font=\scriptsize] {$\mathbf{p}$};
    
    \draw[standard-arrow] (edgescore.east) -| ([xshift=2.5cm]gene.south) node[pos=0.25, above, font=\scriptsize] {$s_{edge}$};
    
    \draw[standard-arrow] (loader.south) -- ++(0,-2.7cm) -| ([xshift=-2.5cm]gene.south) node[pos=0.03, above, font=\scriptsize] {$S_g$};
    
    \node[highbox, retrieval-color, right=2.0 cm of gene, yshift=0cm] (highbox) {
      \scriptsize \textbf{Final}\\
      \scriptsize \textbf{Patient}\\
      \scriptsize \textbf{Subgraph}
    };
    \draw[dashed-arrow] (gene.east) -- (highbox.west);
    \draw[dashed-arrow] (edgescore.east) -- ++(23.5cm,0);
    \draw[dashed-arrow] (scaleddot.east) -- ++(23.5cm,0);
    
    \begin{pgfonlayer}{background}
    \node[dashedbox, fit=(gat)(scaleddot)(gene)(edgescore), 
          label={[anchor=north west]north west:\scriptsize\textbf{\methodName{}}}] {};
    \end{pgfonlayer}
    \end{tikzpicture}
    \caption{\textbf{RareNet} - Based on the patient phenotypes we sample a phenotype subgraph (SG) comprising nodes up to $m$-hops away from the phenotypes and extract the included gene list ($S_g$). This SG is fed into the GAT layers, which compute updated embeddings ($h^{(l)}_i$ and $\alpha_{ij}$) that yield the patient representation ($\mathbf{p}$) and edge scores ($s_{edge}$). These features, along with $h^{(l)}_i$ and $S_g$, are combined in the Gene Scorer to produce gene scores ($s_{gene}$), and are finally consolidated in the final patient subgraph.}
    \label{fig:training}
    \end{figure*}
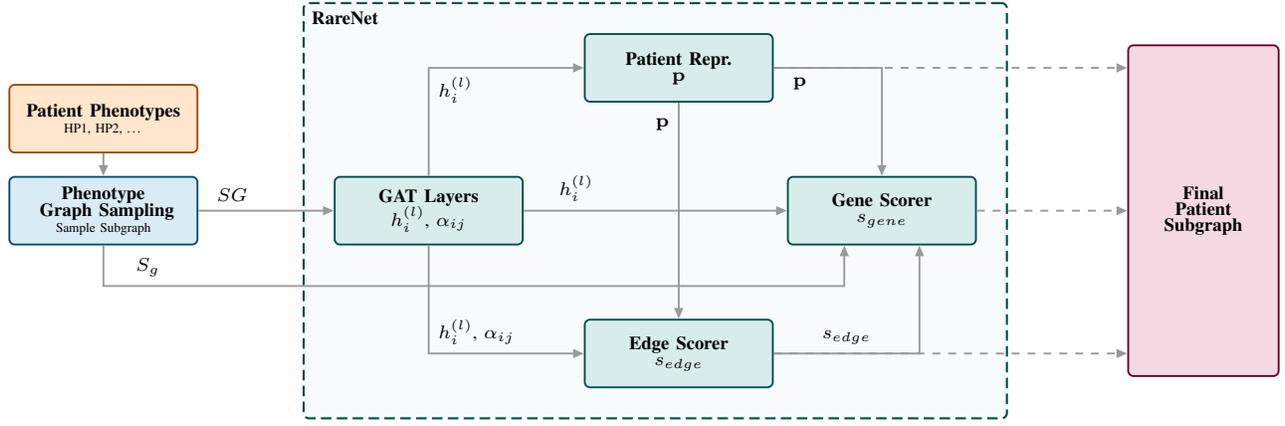 
Our proposed method, RareNet, predicts causal genes without relying on expert-curated candidate gene lists and extracts a compact, informative subgraph of patient-relevant genes from a large-scale knowledge graph. As illustrated in \Cref{fig:training}, RareNet first samples a subgraph centered on a patient's phenotypes, then learns contextual node representations and computes relevance scores across the subgraph. These representations are used to construct a patient-specific profile, assess the importance of graph edges, and prioritize candidate genes, ultimately producing an interpretable subgraph and a ranked gene list for each patient. To train the model, we jointly optimize two objectives: a subgraph loss, which encourages the extraction of informative subgraphs, and a gene loss, which ensures the accurate prioritization of the causal gene among all candidates. We hypothesize that the joint optimization of a subgraph loss and a gene loss will enable the model to effectively extract informative subgraphs while ensuring accurate prioritization of the true causal gene. 
\subsection{Model Architecture}
Let rare disease knowledge graph be defined as $\mathcal{G} = (V, E)$, where $V$ is a set of different types of nodes, including phenotypes, diseases, and genes, and $E \subseteq \{\{u, v\}|u,v \in V\}$ is a set of edges between $u$ and $v$. Each node $v \in V$ is initialized with a randomly assigned learnable embedding $\mathbf{h}_v^{(0)}$.

\paragraph{Phenotype Graph Sampling.}  
Given a set of patient phenotypes $\mathcal{P} \subseteq V$, we begin by extracting an $m$-hop neighborhood from the knowledge graph, centered on the phenotypes. This produces a phenotype-centered subgraph, $\mathcal{N}_m(\mathcal{P})$, which contains all nodes reachable within $m$ hops from any phenotype in $\mathcal{P}$. This sampled subgraph forms the basis for downstream representation learning and gene prioritization.

\paragraph{GNN Processing.}  
The extracted subgraph is then processed by a stack of Graph Attention Network (GAT) layers~\cite{brody2022how}, which iteratively update node representations by aggregating information from their neighbors, weighted by learned attention coefficients. Let $\mathbf{h}_i^{(l)}$ denote the embedding of node $i$ at layer $l$, and $\alpha_{ij}^{(k)}$ the attention coefficient for edge $(i,j)$ from head $k$. At each layer, node embeddings are refined as
\begin{equation}
\mathbf{h}_i^{(l+1)} \gets \mathbf{W}_{\text{proj}} \Biggl( \big\Vert_{k=1}^{H} \sigma\!\Biggl( \sum_{j\in \{i\}\cup\mathcal{N}_1(i)} \alpha_{ij}^{(k)}\, \mathbf{W}^{(k)}\, \mathbf{h}_j^{(l)} \Biggr) \Biggr),
\end{equation}
where $\mathcal{N}_1(i)$ denotes the one-hop neighborhood of node $i$, $H$ is the number of attention heads, $\mathbf{W}^{(k)}$ and $\mathbf{W}_{\text{proj}}$ are learnable weight matrices, and $\sigma$ is a non-linear activation function. After $L$ layers, the final node embeddings $\mathbf{h}_i^{(L)}$ and the learned attention scores $\alpha_{ij}^{(k)}$ serve as the basis for constructing the patient representation and for subsequent edge and gene scoring, which are described in the remainder of this section.

\paragraph{Patient Representation.} We represent each patient’s phenotypic profile with a vector $\mathbf{p}$, obtained by aggregating the embeddings of all phenotype nodes in the subgraph using scaled dot-product pooling:
\begin{equation}
    \mathbf{p} = \sum_{k \in \mathcal{P}} 
    \mathrm{softmax}_k \!\Bigl(\frac{\mathbf{h}_k^\top \mathbf{q}}{\sqrt{d}}\Bigr) \mathbf{h}_k,
\end{equation}
where $\mathbf{q}$ is a learnable query vector and $d$ is the embedding dimension. This aggregated representation provides a compact summary of the patient's observed phenotypes and serves as a key input for both edge and gene scoring.

\paragraph{Edge Scoring.}
To quantify edge importance, we compute a patient-specific score for each edge $S_{\mathrm{edge}}(s,t)$ by concatenating four sources of information: 
(i) a projection of the aggregated multi‐head attention vectors across all layers \(\mathrm{Proj}(\mathbf{A}_{(s,t)})\) with \(\mathbf{A}_{(s,t)}\) being the concatenation of all attention vectors over all layers and heads, which captures the message‐passing dynamics in the graph,
(ii) the patient representation, $\mathbf{p}$, which personalizes the score; (iii) the absolute feature difference, $|\mathbf{h}_s - \mathbf{h}_t|$, measuring node dissimilarity; and (iv) the element-wise product, $\mathbf{h}_s \odot \mathbf{h}_t$, capturing feature interaction. The resulting feature vector is passed through a multi-layer perceptron, $\phi$, yielding the final edge score
\begin{equation}
s_{\mathrm{edge}}(s,t) = \phi\left( \left[ \mathrm{Proj}\big(\mathbf{A}_{(s,t)}\big),\; \mathbf{p},\; |\mathbf{h}_s - \mathbf{h}_t|,\; \mathbf{h}_s \odot \mathbf{h}_t \right] \right).
\end{equation}
This design integrates both global structural information and patient-specific context, aiming for robust and interpretable edge relevance estimation.

\paragraph{Gene Scoring.} To prioritize candidate genes, we assign each gene node $g$ a score reflecting its relevance to the patient. The base score is the cosine similarity between the patient representation $\mathbf{p}$ and the gene embedding $\mathbf{h}_g$, further adjusted by aggregating edge scores from all edges $(u, g)$ entering $g$ 
\begin{multline}
    \mathrm{s}_{\mathrm{gene}}(g)
    =
    \frac{\mathbf{p}^\top\,\mathbf{h}_g}{\|\mathbf{p}\|\;\|\mathbf{h}_g\|} \\
    - 
    \lambda \Biggl[1 - \max\Bigl(0, \min\bigl(\mathrm{avg}\{ s_{\mathrm{edge}}(u,g) \}, 1\bigr)\Bigr)\Biggr],
\end{multline}
where $\max(0,\min(x,1))$ clamps $x$ to $[0,1]$. This formulation encourages both the inclusion of the causal gene in the final subgraph by penalizing genes with weak supporting evidence, and enhances the overall accuracy of candidate gene ranking. The resulting gene scores are then used to rank candidate genes for downstream evaluation.

\subsection{Loss Functions and Training Objective}
The training objective comprises two loss terms that are jointly optimized: a subgraph loss, which encourages the extraction of concise, phenotype–gene–enriched subgraphs, and a gene loss, which ensures accurate prioritization of the true causal gene. These are coupled in a teacher–student setup, where the edge scorer guides the gene scorer via a penalty mechanism. The subgraph loss contrasts positive edges, those linking phenotype nodes to the causal gene, against negative edges sampled from the $m$-hop neighborhood. To enable weak supervision in the vast knowledge graph, we extract positive edges from all phenotype-to-causal gene paths and sample $k$ times as many negatives from the remaining edges. Let \(e_i\) denote the score of edge $i$, and let \(\mathcal{E}_{+}\) and \(\mathcal{E}_{-}\) denote the sets of positive and negative edges, respectively. The subgraph loss is given by
\begin{multline}
\mathcal{L}_{\mathrm{sub}} = \frac{1}{|\mathcal{E}_{+}||\mathcal{E}_{-}|}
\sum_{e_{+}\in \mathcal{E}_{+}} \sum_{e_{-}\in \mathcal{E}_{-}} 
\mathrm{ReLU}\Bigl(\gamma - (e_{+} - e_{-})\Bigr) \\
+ \lambda_{1} \|\mathbf{e}\|_{1} 
+ \lambda_{2} \|\mathbf{e}\|_{2}^{2} 
+ \lambda_{\mathrm{sp}}\, \mathcal{R}_{\mathrm{sp}}(\mathbf{e})
\end{multline}
where \(\gamma\) is the margin hyperparameter and \(\mathbf{e}\) is the vector of all edge scores.
In addition to the margin loss, we add L1 and L2 regularization to control the overall magnitude of \(\mathbf{e}\), preventing edge scores from becoming excessively large and undermining thresholding or sparsity during subgraph extraction. We also add a sparsity penalty, \(\lambda_{\mathrm{sp}}\, \mathcal{R}_{\mathrm{sp}}(\mathbf{e})\), to encourage a compact subgraph by penalizing edge scores exceeding a threshold. This ensures only high-confidence edges remain. The second term, the gene loss, optimizes the ranking of the causal gene relative to hard negatives, inspired by the Shepherd gene loss~\cite{shepherd}. Here, hard negatives are defined as non-causal genes with predicted scores above a threshold $t$, i.e., genes that the model assigns high scores and might confuse with the true gene. Let $\{g_{1},\dots,g_{m}\}$ be the genes in the subgraph with scores $\{s_{1},\dots,s_{m}\}$, and let $g_{\mathrm{true}}$ be the causal gene with score $s_{\mathrm{true}}$. The gene loss is formulated as
\begin{multline}
\mathcal{L}_{\mathrm{gene}} =
\frac{1}{\alpha} \log \Bigl( 1 + \exp\bigl[-\alpha (s_{\mathrm{true}} - t)\bigr] \Bigr) \\
+ \frac{1}{\beta} \log \Bigl( 1 + \sum_{n\in\mathrm{HN}} \exp\bigl[\beta (s_n - t)\bigr] \Bigr)
\end{multline}
where $\mathrm{HN}$ denotes the set of hard-negative genes (i.e., $s_n > t$), and $\alpha$, $\beta$ are weighting factors. When the causal gene is absent, the loss penalizes large positive scores for all candidates. The overall training loss combines both terms, jointly training the student and teacher components.

\begin{algorithm}[t]
\caption{Patient Graph Extraction via Edge Scoring}
\label{alg:subgraph}
\begin{algorithmic}[1]
\State \textbf{Input:} Active phenotypes \(\mathcal{P}\), edge \& gene scores \((\mathbf{s}_{\text{edge}},  \mathbf{s}_{\text{gene}})\), gene indices \( \mathcal{G} \), thresholds \( (\tau_{\text{edge}}, \tau_{\text{gene}}) \), Top-k limits \( (k_1, k_2) \), Edge index \(\mathbf{E}\), Number of hops \( m \)
\State \textbf{Output:} Extracted subgraph nodes \(\mathcal{S}\)

\State \textbf{Initialize:}
\State \quad \( \mathcal{S} \gets \mathcal{P} \) \textit{(Final collected subgraph)}
\State \quad \( \mathcal{F}_0 \gets \mathcal{P} \) \textit{(Frontier for expansion)}

\For{\( h = 1 \) to \( m-1 \)}
    \State \( \mathcal{F}_h \gets \emptyset \) \textit{(Initialize next frontier)}
    \For{\textbf{each} node \( v \in \mathcal{F}_{h-1} \)}
        \State Collect neighboring edges \( (v,u) \in \mathbf{E} \) and corresponding scores \( \mathbf{s}_{\text{edge}}(v,u) \)
        \State Select edges where \( \mathbf{s}_{\text{edge}}(v,u) \geq \tau_{\text{edge}} \)
        \If{more than \( k_1 \) edges are selected}
            \State Retain only top \( k_1 \) edges by score
        \EndIf
        \State Add target nodes \( u \) of selected edges to \( \mathcal{F}_{\text{h}} \)
    \EndFor
    \State \( \mathcal{S} \gets \mathcal{S} \cup \mathcal{F}_h \) \textit{(Expand collected subgraph)}
\EndFor
\State \( \mathcal{F}_m \gets \emptyset \) \textit{(Initialize final frontier)}
\For{\textbf{each} node \( w \in \mathcal{F}_{m-1} \)}
    \State Collect neighboring edges \( (w, g) \) where \( g \in \mathcal{G} \)
    \State Select edges where \( \mathbf{s}_{\text{edge}}(w,g) \geq \tau_{\text{edge}} \) and \( \mathbf{s}_{\text{gene}}(g) \geq \tau_{\text{gene}} \)
    \If{more than \( k_2 \) genes are selected}
        \State Retain only top \( k_2 \) edges by score
    \EndIf
    \State Add selected genes \( g \) to \( \mathcal{F}_m \)
\EndFor
\State \( \mathcal{S} \gets \mathcal{S} \cup \mathcal{F}_m \) \textit{(Expand collected subgraph)}

\State \textbf{Return} \( \mathcal{S} \)
\end{algorithmic}
\end{algorithm}

\subsection{Patient Graph Extraction}
The previous stage produces edge and gene scores for the phenotype-centered subgraph. Retaining all scored nodes and edges would yield large, noisy structures, so we apply an $m$-hop filtering strategy to construct a focused, patient-specific subgraph. Starting from the active phenotypes, we iteratively expand for $m-1$ hops, retaining only neighboring nodes connected by edges above a score threshold. In the final ($m$-th) hop, expansion is limited to gene nodes meeting both edge and gene score thresholds. This ensures the resulting subgraph is concise and enriched for likely causal genes, supporting interpretable ranking and integration with other methods. The full procedure is given in Algorithm~\ref{alg:subgraph}.

\subsection{Enhancing Existing Methods}
RareNet can also be used to enhance other gene prioritization models, either by supplying a compact gene list for pre-processing or by re-ranking candidates in a post-processing step. In the latter setup, we normalize the scores produced by the external method for each patient (e.g., using min-max scaling), and then boost the score of each gene that appears in the RareNet subgraph:
\begin{equation}
    \mathrm{score}_{\text{final}}(g) = \widetilde{s}_g + \delta \cdot \mathbf{1}_{\{g \in \mathcal{S}\}},
\end{equation}
where $\widetilde{s}_g$ is the normalized base score, $\mathcal{S}$ is the set of genes identified by RareNet, and $\delta$ controls the boost magnitude. Genes are then re-ranked by $\mathrm{score}_{\text{final}}(g)$. This lightweight fusion incorporates RareNet's patient-specific graph signal, without retraining the base model.

\section{Experimental Setup}
\label{sec:exp_results}
\begin{table}[t]
\centering
\caption{Summary statistics comparing the simulated and MyGene2 datasets. \#Pts = number of patients, \#Phenos = number of unique phenotypes, Avg/Pt = average phenotypes per patient, \#Genes = number of unique causal genes.}
\begin{tabular}{lrrrr}
\toprule
\textbf{Dataset} & \textbf{\#Pts} & \textbf{\#Phenos} & \textbf{Avg/Pt} & \textbf{\#Genes} \\
\midrule
MyGene2 & 146 & 294 & 7.9 & 48 \\
Simulated & 6400 & 4107 & 18.7 & 499 \\
\bottomrule
\end{tabular}
\vspace{2mm}
\label{tab:datasets_comparison}
\end{table}

\paragraph{Datasets.} 
We conduct our experiments on two datasets: a real-world dataset, MyGene2 \cite{uwcmg_mygene2}, and a simulated dataset \citep{alsentzer2023simulation}. We utilize the knowledge graph from \citet{chandak2023building}, following the same preprocessing steps as in \citet{shepherd}. It comprises $105,220$ nodes and $1,095,469$ edges. The Simulated Patient dataset comprises $42$K patients ($36$K for training and $6.4$K for testing, Table \ref{tab:datasets_comparison}). The test set of Simulated Patients is designed to partially exclude diseases and genes present in the training set, ensuring that it primarily contains unseen cases. This setup provides a robust evaluation of the model's ability to generalize to novel data. We further evaluate a second dataset, the MyGene2 real-world public dataset with a total number of 146 patients, and a detailed description and preprocessing steps can be found in \cite{shepherd}. 

\paragraph{Baselines.}
We compare our results with recent models, including Shepherd \cite{shepherd}, PhenoApt \cite{{cui2020conan}}, and Amelie \cite{birgmeier2020amelie}. While Shepherd and Amelie work with an expert-curated candidate gene list, our method does not need this input. In order to facilitate a fair comparison, we consider all genes in 2-hops from each phenotype as candidate genes, which is $\approx$ 6k genes, the same as for our method. PhenoApt operates without requiring an explicit candidate gene list, as it retrieves this information internally. We used the Amelie and PhenoApt API, as well as the official code and checkpoints provided by Shepherd. 
We investigate whether RareNet's extracted patient subgraphs can be used to improve the performance of existing causal gene prioritization methods. Instead of running methods like Shepherd and Amelie on the full 2-hop phenotype subgraph, we restrict their candidate gene to ones obtained from the compact, patient-specific subgraphs generated by \methodName{}. We refer to this enhanced setup as + RN. For PhenoApt, since it does not require a candidate gene list, we applied the post-processing filter described in Section: Enhancing Existing Methods.
We exclude AI-MARRVEL~\cite{mao2024ai} from direct benchmarking, as it requires both phenotype and patient-specific genotype data (VCF files), which falls outside the phenotype-only scope assumed by RareNet.

\paragraph{Subset Selection for Evaluation.}
Due to the time-consuming nature of Amelie's API (3 minutes per patient for $\approx$6K genes) making full-scale evaluation infeasible, we evaluated it on a randomly selected subset of the Simulated Patients test set (2370 patients). For consistency, all other methods were evaluated under two settings: on the full Simulated Patients dataset and on the same subset used for Amelie called Simulated Patients Subset.

\paragraph{Evaluation Metrics.}
We evaluate model performance using two standard ranking metrics: (1) Hit@k, which measures the proportion of patients for whom the true causal gene appears among the top-$k$ predictions (we report Hit@1, Hit@5, and Hit@10); and (2) Mean Reciprocal Rank (MRR), which captures the average inverse rank of the causal gene across all patients, giving higher weight to top-ranked hits. All values are shown in percentages.

\paragraph{Implementation Details.}
 Our model is implemented in PyTorch (v2.0.1) \cite{paszke2017automatic} and PyTorch Geometric (v2.5.3) \cite{fey2019fast}. Node embeddings are randomly initialized and learned during training. The architecture employs a hidden dimension of 128, an output dimension of 64, 4 attention heads, and 3 GATv2 layers \cite{brody2022how} (referred to as GAT for simplicity), resulting in approximately 130K parameters. We optimize the model using Adam Optimizer with a learning rate of \(1 \times 10^{-4}\) and a step scheduler (step size 10, \(\eta=0.5\)). For patient graph extraction (\textit{cf} Algorithm~\ref{alg:subgraph}), we use the following settings: \(m=2\) hops, top-\(k\) limits \(k_1=5\) for edge selection and \(k_2=2\) for gene selection, \(\tau_{\text{edge}}\) is set as the 80th percentile of edge scores, and \(\tau_{\text{gene}}=0.5\). For the subgraph loss, we set \(\lambda_1=0.15\), \(\lambda_2=0.15\), \(\lambda_{\mathrm{sp}}=0.25\), and $\gamma=0.5$. For the gene loss, we use \(\alpha=2\), \(\beta=40\), and \(t=0.5\). For the post-processing fusion strategy, 
 we used thresholds of $\delta = 0.6$ for Simulated Patents and $0.1$ for MyGene2. We ran our method three times to compute the standard deviation. All experiments were conducted on a single NVIDIA A40 GPU. 
\section{Results and Discussions}
\subsection{Main Results}
\begin{table}[t]
\centering
\caption{Performance comparison on Simulated and MyGene2 datasets using standard graphs versus RareNet-derived patient subgraphs (denoted as +RN). Best in each method group is \underline{underlined}; overall bests are \textbf{bold}.}
\setlength{\tabcolsep}{1mm}
\small
\begin{tabular}{llrrrr}
\toprule
\textbf{Dataset} & \textbf{Method} & \textbf{Hits@1} & \textbf{Hits@5} & \textbf{Hits@10} & \textbf{MRR} \\
\midrule
\multicolumn{6}{l}{\textit{Simulated Patients Subset}} \\
& Shepherd           & 13.3 & 27.2 & 35.1 & 20.8 \\
& \quad + RN         & \underline{15.4} & \underline{37.8} & \underline{51.0} & \underline{28.3} \\
\cmidrule(l){2-6}
& Amelie             & 4.4 & 12.7 & 18.1 & 8.9 \\
& \quad + RN         & \underline{10.5} & \underline{28.0} & \underline{40.2} & \underline{19.7} \\
\cmidrule(l){2-6}
& PhenoApt           & 8.8 & 17.0 & 22.7 & 13.3 \\
& \quad + RN         & \underline{11.0} & \underline{24.9} & \underline{34.3} & \underline{17.7} \\
\cmidrule(l){2-6}
& RareNet            & \textbf{27.0 ± 1.0} & \textbf{52.3 ± 1.3} & \textbf{61.0 ± 2.2} & \textbf{37.3 ± 0.4} \\
\midrule
\multicolumn{6}{l}{\textit{MyGene2}} \\
& Shepherd           & 10.3 & 20.6 & 25.7 & 17.0 \\
& \quad + RN         & \underline{17.1} & \underline{43.2} & \underline{50.0} & \underline{30.5} \\
\cmidrule(l){2-6}
& Amelie             & 28.5 & 46.5 & \underline{59.0} & 37.6 \\
& \quad + RN         & \underline{32.6} & \underline{50.6} & 55.5 & \underline{40.2} \\
\cmidrule(l){2-6}
& PhenoApt           & 41.8 & \underline{61.6} & \underline{67.8} & \underline{50.5} \\
& \quad + RN         & \textbf{43.2} & \textbf{61.6} & \textbf{67.8} & \textbf{51.2} \\
\cmidrule(l){2-6}
& RareNet            & 13.0 ± 5.8 & 27.4 ± 3.8 & 35.2 ± 7.2 & 20.2 ± 2.5 \\
\bottomrule
\end{tabular}
\label{tab:shepherd_vs_ours_plus_shepherd}
\end{table}

\begin{table}[t]
\centering
\caption{Causal gene prediction performance of Shepherd, PhenoApt, and RareNet on two simulated datasets. All values are shown in percentages. Bold indicates the best performance.}
\setlength{\tabcolsep}{1mm}
\small
\begin{tabular}{llrrrr}
\toprule
\textbf{Dataset} & \textbf{Method} & \textbf{Hits@1} & \textbf{Hits@5} & \textbf{Hits@10} & \textbf{MRR} \\
\midrule
\multicolumn{6}{l}{\textit{Simulated Patients Set}} \\
& Shepherd  & 13.0 & 27.0 & 35.7 & 20.8 \\
& PhenoApt  & 8.8  & 17.8 & 23.3 & 13.1 \\
& RareNet   & \textbf{27.2 ± 1.0} & \textbf{51.3 ± 1.4} & \textbf{60.6 ± 2.2} & \textbf{37.3 ± 0.4} \\
\midrule
\multicolumn{6}{l}{\textit{Simulated Patients (Mixed)}} \\
& Shepherd  & 13.1 & 28.8 & 37.7 & 21.5 \\
& PhenoApt  & 9.0  & 17.7 & 23.5 & 13.7 \\
& RareNet   & \textbf{57.4 ± 1.0} & \textbf{82.7 ± 0.4} & \textbf{89.2 ± 0.5} & \textbf{68.4 ± 0.9} \\
\bottomrule
\end{tabular}
\label{tab:shepherd_vs_ours}
\end{table}

\begin{table}[t]
\centering
\caption{Performance of Shepherd, Amelie, and PhenoApt on the MyGene2 patients where the causal gene is included in the RareNet subgraph. Results are shown without and with RareNet patient subgraphs (denoted as +RN).}
\setlength{\tabcolsep}{1mm}
\small
\begin{tabular}{lrrrr}
\toprule
\textbf{Method} & \textbf{Hits@1} & \textbf{Hits@5} & \textbf{Hits@10} & \textbf{MRR} \\
\midrule
Shepherd        & 17.1 & 30.5 & 36.6 & 25.6 \\
\quad + RN      & \textbf{30.5} & \textbf{76.8} & \textbf{89.0} & \textbf{47.2} \\
\midrule
Amelie          & 40.2 & 54.8 & 72.8 & 49.8 \\
\quad + RN      & \textbf{57.3} & \textbf{89.0} & \textbf{97.6} & \textbf{70.6} \\
\midrule
PhenoApt        & 67.1 & 79.3 & 82.9 & 72.8 \\
\quad + RN      & \textbf{74.4} & \textbf{84.1} & \textbf{86.6} & \textbf{79.5} \\
\bottomrule
\end{tabular}
\label{tab:amelie_phenoapt_rarenet_82}
\end{table}

\begin{table}[t]
\centering
\caption{Comparison of RareNet and RareNet (Pretrained) on Simulated Patients. Bold indicates the best performance per column.}
\setlength{\tabcolsep}{1mm}
\small
\begin{tabular}{lrrrr}
\toprule
\textbf{Method} & \textbf{Hits@1} & \textbf{Hits@5} & \textbf{Hits@10} & \textbf{MRR} \\
\midrule
RareNet              & 26.2 & \textbf{53.1} & \textbf{63.2} & \textbf{37.6} \\
RareNet (Pretrained) & \textbf{28.6} & 49.9 & 59.1 & 37.6 \\
\bottomrule
\end{tabular}
\label{tab:ours_vs_pretrained}
\end{table}

\begin{figure}[t]
    \centering
    \includegraphics[width=\columnwidth]{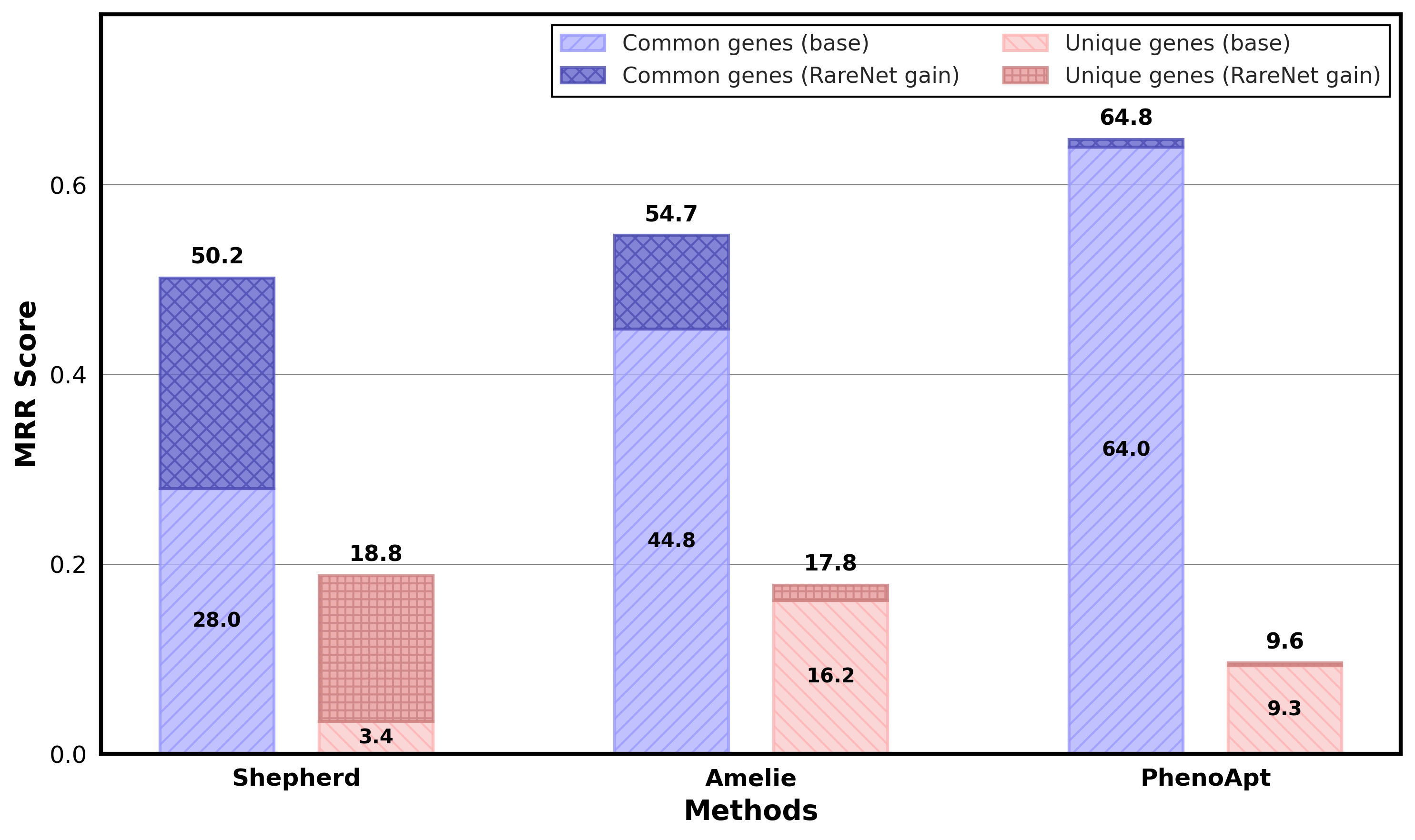}
    \caption{MRR performance of the different methods comparing patients with common causal genes versus unique gene cases.}
    \label{fig:mrr_rarenet_gain}
\end{figure}
\Cref{tab:shepherd_vs_ours_plus_shepherd} summarizes the performance of RareNet compared to existing causal gene prioritization methods on both the Simulated Patients Subset and MyGene2 datasets. In the Simulated Patients Subset, RareNet achieves the best results across all metrics, substantially outperforming all baselines (e.g., Hits@1 of 27.0\% vs. 15.4\% for Shepherd + RN). Compared to the standard Shepherd, RareNet more than doubles Hits@1, demonstrating the strength of patient-specific subgraph modeling. Other methods, such as PhenoApt and Amelie, also perform significantly worse than RareNet, with PhenoApt reaching an MRR of 13.3\% and Amelie only 8.9\%. Their weaker performance is largely due to the higher diversity of causal genes in the simulated dataset, which challenges models that rely heavily on prior gene-frequency bias or limited gene-phenotype associations (\textit{cf.} \Cref{tab:shepherd_vs_ours_plus_shepherd}).

\noindent On the MyGene2 dataset, RareNet achieves moderate performance (MRR = 20.2\%), while methods like PhenoApt and Amelie report higher scores-PhenoApt with Hits@1 = 41.8\%, MRR = 50.5\%; Amelie with MRR = 37.6\%. However, these scores are largely driven by a small set of highly overrepresented causal genes- PIEZO2, NALCN, DHODH, and FBN2-which collectively account for around 75\% of the cases. While MyGene2 contains 146 patients, it includes only 48 unique causal genes. For example, PIEZO2 appears in 36 cases, and both NALCN and SF3B4 appear in 15. PhenoApt achieves high accuracy on these common genes (e.g., PIEZO2 MRR = 89.6\%), but performance drops to near zero on rarer ones like TNNI2 and MYH3. This is reflected in the average MRR gap: 64.0\% for common-gene cases vs. only 9.3\% for rare ones (\textit{cf.} \Cref{fig:mrr_rarenet_gain}).

\noindent This challenge is further highlighted in the simulated test cohort, which includes a more diverse set of phenotypes and causal genes (\textit{cf.} \Cref{tab:datasets_comparison}). Frequent MyGene2 genes are largely absent-e.g., PIEZO2 is not present, NALCN appears in only 4 cases, and SF3B4 in 6. Both Amelie and PhenoApt perform significantly worse on these genes in the simulated setting-e.g., NALCN MRR drops to 0.6\% (PhenoApt) and 9.9\% (Amelie); for SF3B4, to 7.7\% and 22.5\%, respectively. This decline is due to both increased phenotype noise and gene diversity, which better reflect real-world clinical inputs derived from unstructured medical notes.

\noindent In contrast, RareNet and Shepherd maintain more stable performance across both datasets. On the simulated subset, RareNet achieves 27.0\% Hits@1 and 37.3\% MRR, while Shepherd scores 13.3\% Hits@1 and 20.8\% MRR. On MyGene2, RareNet reports 13.0\% Hits@1 and 20.2\% MRR, while Shepherd achieves 10.3\% and 17.0\%, respectively. This consistency stems from training on a knowledge graph that treats common and rare genes equally, reducing reliance on frequency-based shortcuts and enhancing robustness in realistic clinical settings.

\subsection{Enhancing Existing Methods} 
As shown in \Cref{tab:shepherd_vs_ours_plus_shepherd}, integrating RareNet-derived subgraphs as candidate gene sets consistently improves performance across both datasets. On the Simulated Patients Subset, Shepherd’s MRR increases from 20.8\% to 28.3\%, Amelie improves from 8.9\% to 19.7\%, and PhenoApt from 13.3\% to 17.7\%. Gains are even more pronounced on the MyGene2 dataset: Shepherd’s MRR nearly doubles from 17.0\% to 30.5\%, Amelie improves from 37.6\% to 40.2\%, and PhenoApt from 50.5\% to 51.2\%.

\noindent To better understand these gains, we conducted an ablation study on the MyGene2 dataset (\textit{cf.} \Cref{fig:mrr_rarenet_gain}), comparing performance separately on patients with common versus unique causal genes. RareNet-derived subgraphs substantially boosted Shepherd and Amelie’s performance on common genes-Shepherd’s MRR increased from 28.0\% to 50.2\%, and Amelie’s from 44.8\% to 54.7\%. For unique gene cases, the improvements were more modest. For instance, Shepherd’s MRR increased slightly from 18.8\% to 22.2\% and Amelie’s from 17.8\% to 19.4\%. These findings indicate that RareNet can mitigate gene-frequency bias for certain models by narrowing the ranking to a more patient-relevant subset of genes.

\subsection{Ablation Studies} 
To evaluate the performance of each method under different levels of gene exposure during training, we created two versions of the Simulated Patients dataset and trained our model separately on each. We consider two data splits: (1) the curated split used by Shepherd-where causative genes in the test set are not seen during training-referred to simply as Simulated Patients; and (2) a Mixed (Random) Split, which is a standard train/test split of our full 42K-patient dataset, referred to as Simulated Patients (Mixed). Evaluation of Amelie in this setting was not feasible. 

\noindent \methodName{} significantly outperforms competitive methods on the mixed simulated patients split, for example, achieving a 4.3× higher Hit@1 score than Shepherd (57.4\% vs. 13.1\%). This demonstrates that when causative genes are seen during training, the model’s performance improves substantially compared to evaluation on entirely unseen genes (\textit{cf.} \Cref{tab:shepherd_vs_ours}, Simulated Patients). In contrast, Shepherd and PhenoApt do not exhibit improved performance on the mixed setting, likely because they had already been exposed to all genes during pretraining on the full knowledge graph. 
\noindent Furthermore, most KG-based models, including Shepherd, require extensive pretraining of knowledge graph embeddings on link prediction tasks to achieve competitive performance. As shown in \Cref{tab:ours_vs_pretrained}, our method does not rely on this step and achieves comparable performance with and without pretraining. This highlights the robustness and effectiveness of our end-to-end training approach. Note that the causal gene may or may not be included in the patient-specific subgraph extracted from the knowledge graph. Since the model only ranks genes within this subgraph, the exclusion of the causal gene prevents it from being evaluated, which negatively impacts metrics like Hit@k and MRR, as these cases are treated as incorrect predictions. To quantify the impact of this issue, we evaluated the percentage of true causal genes that were included in the extracted subgraphs across all datasets. We found inclusion rates of 83. 58\% for Simulated Patients, 88.52\% for Simulated Patients (Mixed), and 56.9\% for MyGene2. 

\noindent We also evaluated the performance of Amelie and PhenoApt separately on 56.9\% of MyGene2 patients for whom the causal gene is included in the RareNet-generated subgraph (\textit{cf.} \Cref{tab:amelie_phenoapt_rarenet_82}). In this subset, where the causal gene appears in the list of candidate genes, both methods show even greater improvements compared to their performance on the full test set (\textit{cf.} \Cref{tab:shepherd_vs_ours_plus_shepherd}), where no filtering based on causal gene presence was applied. 
This aligns with how recent candidate-gene-based methods-such as Shepherd and Amelie - are typically evaluated. Since these methods rank genes from a predefined candidate list, it is typically assumed that the causal gene is present in this list during evaluation - an assumption that does not always hold in real-world scenarios. This further highlights the importance of methods that do not rely on a pre-specified candidate gene list, such as RareNet.
\section{Conclusion}
We present RareNet, a subgraph-based Graph Neural Network that operates without relying on a predefined candidate gene list, addressing a critical gap in causal gene prediction for rare diseases. Our method can function as a standalone predictor or enhance existing approaches by either generating patient-specific candidate gene lists or serving as a post-processing step to improve their results.
Through comprehensive evaluations, we demonstrate competitive performance against state-of-the-art methods and consistent improvements when combined with them. We also raise important questions about current evaluation protocols, benchmarking practices, and the limitations of existing datasets.

\bibliographystyle{IEEEtran}
\bibliography{references}

\begin{thebibliography}{10}
\providecommand{\url}[1]{#1}
\csname url@samestyle\endcsname
\providecommand{\newblock}{\relax}
\providecommand{\bibinfo}[2]{#2}
\providecommand{\BIBentrySTDinterwordspacing}{\spaceskip=0pt\relax}
\providecommand{\BIBentryALTinterwordstretchfactor}{4}
\providecommand{\BIBentryALTinterwordspacing}{\spaceskip=\fontdimen2\font plus
\BIBentryALTinterwordstretchfactor\fontdimen3\font minus \fontdimen4\font\relax}
\providecommand{\BIBforeignlanguage}[2]{{%
\expandafter\ifx\csname l@#1\endcsname\relax
\typeout{** WARNING: IEEEtran.bst: No hyphenation pattern has been}%
\typeout{** loaded for the language `#1'. Using the pattern for}%
\typeout{** the default language instead.}%
\else
\language=\csname l@#1\endcsname
\fi
#2}}
\providecommand{\BIBdecl}{\relax}
\BIBdecl

\bibitem{marwaha2022guide}
S.~Marwaha, J.~W. Knowles, and E.~A. Ashley, ``A guide for the diagnosis of rare and undiagnosed disease: beyond the exome,'' \emph{Genome medicine}, vol.~14, no.~1, p.~23, 2022.

\bibitem{steinhaus2021mutationtaster2021}
R.~Steinhaus, S.~Proft, M.~Schuelke, D.~N. Cooper, J.~M. Schwarz, and D.~Seelow, ``Mutationtaster2021,'' \emph{Nucleic Acids Research}, vol.~49, no.~W1, pp. W446--W451, 2021.

\bibitem{rentzsch2019cadd}
P.~Rentzsch, D.~Witten, G.~M. Cooper, J.~Shendure, and M.~Kircher, ``Cadd: predicting the deleteriousness of variants throughout the human genome,'' \emph{Nucleic acids research}, vol.~47, no.~D1, pp. D886--D894, 2019.

\bibitem{jagadeesh2016m}
K.~A. Jagadeesh, A.~M. Wenger, M.~J. Berger, H.~Guturu, P.~D. Stenson, D.~N. Cooper, J.~A. Bernstein, and G.~Bejerano, ``M-cap eliminates a majority of variants of uncertain significance in clinical exomes at high sensitivity,'' \emph{Nature genetics}, vol.~48, no.~12, pp. 1581--1586, 2016.

\bibitem{yang2015phenolyzer}
H.~Yang, P.~N. Robinson, and K.~Wang, ``Phenolyzer: phenotype-based prioritization of candidate genes for human diseases,'' \emph{Nature methods}, vol.~12, no.~9, pp. 841--843, 2015.

\bibitem{jagadeesh2019phrank}
K.~A. Jagadeesh, J.~Birgmeier, H.~Guturu, C.~A. Deisseroth, A.~M. Wenger, J.~A. Bernstein, and G.~Bejerano, ``Phrank measures phenotype sets similarity to greatly improve mendelian diagnostic disease prioritization,'' \emph{Genetics in Medicine}, vol.~21, no.~2, pp. 464--470, 2019.

\bibitem{peng2021cada}
C.~Peng, S.~Dieck, A.~Schmid, A.~Ahmad, A.~Knaus, M.~Wenzel, L.~Mehnert, B.~Zirn, T.~Haack, S.~Ossowski \emph{et~al.}, ``Cada: phenotype-driven gene prioritization based on a case-enriched knowledge graph,'' \emph{NAR Genomics and Bioinformatics}, vol.~3, no.~3, p. lqab078, 2021.

\bibitem{cui2020conan}
L.~Cui, S.~Biswal, L.~M. Glass, G.~Lever, J.~Sun, and C.~Xiao, ``Conan: complementary pattern augmentation for rare disease detection,'' in \emph{Proceedings of the AAAI conference on artificial intelligence}, vol.~34, no.~01, 2020, pp. 614--621.

\bibitem{10.1093/nar/gkaa1043}
\BIBentryALTinterwordspacing
S.~Köhler, M.~Gargano, N.~Matentzoglu, L.~C. Carmody, D.~Lewis-Smith, N.~A. Vasilevsky, D.~Danis, G.~Balagura, G.~Baynam, A.~M. Brower, T.~J. Callahan, C.~G. Chute, J.~L. Est, P.~D. Galer, S.~Ganesan, M.~Griese, M.~Haimel, J.~Pazmandi, M.~Hanauer, N.~L. Harris, M.~J. Hartnett, M.~Hastreiter, F.~Hauck, Y.~He, T.~Jeske, H.~Kearney, G.~Kindle, C.~Klein, K.~Knoflach, R.~Krause, D.~Lagorce, J.~A. McMurry, J.~A. Miller, M.~C. Munoz-Torres, R.~L. Peters, C.~K. Rapp, A.~M. Rath, S.~A. Rind, A.~Z. Rosenberg, M.~M. Segal, M.~G. Seidel, D.~Smedley, T.~Talmy, Y.~Thomas, S.~A. Wiafe, J.~Xian, Z.~Yüksel, I.~Helbig, C.~J. Mungall, M.~A. Haendel, and P.~N. Robinson, ``The human phenotype ontology in 2021,'' \emph{Nucleic Acids Research}, vol.~49, no.~D1, pp. D1207--D1217, 12 2020. [Online]. Available: \url{https://doi.org/10.1093/nar/gkaa1043}
\BIBentrySTDinterwordspacing

\bibitem{10.1093/nar/gky1151}
\BIBentryALTinterwordspacing
J.~S. Amberger, C.~A. Bocchini, A.~F. Scott, and A.~Hamosh, ``Omim.org: leveraging knowledge across phenotype–gene relationships,'' \emph{Nucleic Acids Research}, vol.~47, no.~D1, pp. D1038--D1043, 11 2018. [Online]. Available: \url{https://doi.org/10.1093/nar/gky1151}
\BIBentrySTDinterwordspacing

\bibitem{weinreich2008orphanet}
S.~S. Weinreich, R.~Mangon, J.~J. Sikkens, M.~E. Teeuw, and M.~C. Cornel, ``\BIBforeignlanguage{Dutch}{Orphanet: een europese database over zeldzame ziekten},'' \emph{\BIBforeignlanguage{Dutch}{Nederlands Tijdschrift voor Geneeskunde}}, vol. 152, no.~9, pp. 518--519, 2008, article in Dutch (English abstract available).

\bibitem{birgmeier2020amelie}
J.~Birgmeier, M.~Haeussler, C.~A. Deisseroth, E.~H. Steinberg, K.~A. Jagadeesh, A.~J. Ratner, H.~Guturu, A.~M. Wenger, M.~E. Diekhans, P.~D. Stenson \emph{et~al.}, ``Amelie speeds mendelian diagnosis by matching patient phenotype and genotype to primary literature,'' \emph{Science Translational Medicine}, vol.~12, no. 544, p. eaau9113, 2020.

\bibitem{shepherd}
E.~Alsentzer, M.~M. Li, S.~N. Kobren, A.~Noori, U.~D. Network, I.~S. Kohane, and M.~Zitnik, ``Few shot learning for phenotype-driven diagnosis of patients with rare genetic diseases,'' \emph{npj Digital Medicine}, vol.~8, no.~1, p. 380, 2025.

\bibitem{smedley2015next}
D.~Smedley, J.~O. Jacobsen, M.~J{\"a}ger, S.~K{\"o}hler, M.~Holtgrewe, M.~Schubach, E.~Siragusa, T.~Zemojtel, O.~J. Buske, N.~L. Washington \emph{et~al.}, ``Next-generation diagnostics and disease-gene discovery with the exomiser,'' \emph{Nature protocols}, vol.~10, no.~12, pp. 2004--2015, 2015.

\bibitem{li2019xrare}
Q.~Li, K.~Zhao, C.~D. Bustamante, X.~Ma, and W.~H. Wong, ``Xrare: a machine learning method jointly modeling phenotypes and genetic evidence for rare disease diagnosis,'' \emph{Genetics in Medicine}, vol.~21, no.~9, pp. 2126--2134, 2019.

\bibitem{mao2024ai}
D.~Mao, C.~Liu, L.~Wang, R.~AI-Ouran, C.~Deisseroth, S.~Pasupuleti, S.~Y. Kim, L.~Li, J.~A. Rosenfeld, L.~Meng \emph{et~al.}, ``Ai-marrvel—a knowledge-driven ai system for diagnosing mendelian disorders,'' \emph{NEJM AI}, vol.~1, no.~5, p. AIoa2300009, 2024.

\bibitem{10.1093/nar/gkz1021}
\BIBentryALTinterwordspacing
J.~Pinero, J.~M. Ramírez-Anguita, J.~Sauch-Pitarch, F.~Ronzano, E.~Centeno, F.~Sanz, and L.~I. Furlong, ``The disgenet knowledge platform for disease genomics: 2019 update,'' \emph{Nucleic Acids Research}, vol.~48, no.~D1, pp. D845--D855, 11 2019. [Online]. Available: \url{https://doi.org/10.1093/nar/gkz1021}
\BIBentrySTDinterwordspacing

\bibitem{drkg2020}
V.~N. Ioannidis, X.~Song, S.~Manchanda, M.~Li, X.~Pan, D.~Zheng, X.~Ning, X.~Zeng, and G.~Karypis, ``Drkg - drug repurposing knowledge graph for covid-19,'' \url{https://github.com/gnn4dr/DRKG/}, 2020.

\bibitem{10.1093/nar/gkac957}
\BIBentryALTinterwordspacing
F.~Feng, F.~Tang, Y.~Gao, D.~Zhu, T.~Li, S.~Yang, Y.~Yao, Y.~Huang, and J.~Liu, ``Genomickb: a knowledge graph for the human genome,'' \emph{Nucleic Acids Research}, vol.~51, no.~D1, pp. D950--D956, 11 2022. [Online]. Available: \url{https://doi.org/10.1093/nar/gkac957}
\BIBentrySTDinterwordspacing

\bibitem{Chandak2022.05.01.489928}
\BIBentryALTinterwordspacing
P.~Chandak, K.~Huang, and M.~Zitnik, ``Building a knowledge graph to enable precision medicine,'' \emph{bioRxiv}, 2022. [Online]. Available: \url{https://www.biorxiv.org/content/early/2022/05/01/2022.05.01.489928}
\BIBentrySTDinterwordspacing

\bibitem{vilela2023biomedical}
J.~Vilela, M.~Asif, A.~R. Marques, J.~X. Santos, C.~Rasga, A.~Vicente, and H.~Martiniano, ``Biomedical knowledge graph embeddings for personalized medicine: Predicting disease-gene associations,'' \emph{Expert Systems}, vol.~40, no.~5, p. e13181, 2023.

\bibitem{ektefaie2023multimodal}
Y.~Ektefaie, G.~Dasoulas, A.~Noori, M.~Farhat, and M.~Zitnik, ``Multimodal learning with graphs,'' \emph{Nature Machine Intelligence}, vol.~5, no.~4, pp. 340--350, 2023.

\bibitem{galkin2023towards}
\BIBentryALTinterwordspacing
M.~Galkin, X.~Yuan, H.~Mostafa, J.~Tang, and Z.~Zhu, ``Towards foundation models for knowledge graph reasoning,'' in \emph{NeurIPS 2023 Workshop: New Frontiers in Graph Learning}, 2023. [Online]. Available: \url{https://openreview.net/forum?id=LzMWMJlxHg}
\BIBentrySTDinterwordspacing

\bibitem{alsentzer2020subgraph}
E.~Alsentzer, S.~Finlayson, M.~Li, and M.~Zitnik, ``Subgraph neural networks,'' \emph{Advances in Neural Information Processing Systems}, vol.~33, pp. 8017--8029, 2020.

\bibitem{sun2019pullnet}
H.~Sun, T.~Bedrax-Weiss, and W.~Cohen, ``Pullnet: Open domain question answering with iterative retrieval on knowledge bases and text,'' in \emph{Proceedings of the 2019 Conference on Empirical Methods in Natural Language Processing and the 9th International Joint Conference on Natural Language Processing (EMNLP-IJCNLP)}, 2019, pp. 2380--2390.

\bibitem{shen2022improving}
Y.~Shen, X.~Liu, C.-W. Ju, J.~Yan, J.~Yi, Z.~Lin, and H.~Guan, ``Improving subgraph representation learning via multi-view augmentation,'' \emph{arXiv preprint arXiv:2205.13038}, 2022.

\bibitem{alsentzer2023simulation}
E.~Alsentzer, S.~G. Finlayson, M.~M. Li, U.~D. Network, S.~N. Kobren, and I.~S. Kohane, ``Simulation of undiagnosed patients with novel genetic conditions,'' \emph{Nature Communications}, vol.~14, no.~1, p. 6403, 2023.

\bibitem{mavromatis2024gnnraggraphneuralretrieval}
\BIBentryALTinterwordspacing
C.~Mavromatis and G.~Karypis, ``Gnn-rag: Graph neural retrieval for large language model reasoning,'' 2024. [Online]. Available: \url{https://arxiv.org/abs/2405.20139}
\BIBentrySTDinterwordspacing

\bibitem{NEURIPS2024_efaf1c97}
\BIBentryALTinterwordspacing
X.~He, Y.~Tian, Y.~Sun, N.~V. Chawla, T.~Laurent, Y.~LeCun, X.~Bresson, and B.~Hooi, ``G-retriever: Retrieval-augmented generation for textual graph understanding and question answering,'' in \emph{Advances in Neural Information Processing Systems}, A.~Globerson, L.~Mackey, D.~Belgrave, A.~Fan, U.~Paquet, J.~Tomczak, and C.~Zhang, Eds., vol.~37.\hskip 1em plus 0.5em minus 0.4em\relax Curran Associates, Inc., 2024, pp. 132\,876--132\,907. [Online]. Available: \url{https://proceedings.neurips.cc/paper_files/paper/2024/file/efaf1c9726648c8ba363a5c927440529-Paper-Conference.pdf}
\BIBentrySTDinterwordspacing

\bibitem{wen2024mindmapknowledgegraphprompting}
\BIBentryALTinterwordspacing
Y.~Wen, Z.~Wang, and J.~Sun, ``Mindmap: Knowledge graph prompting sparks graph of thoughts in large language models,'' 2024. [Online]. Available: \url{https://arxiv.org/abs/2308.09729}
\BIBentrySTDinterwordspacing

\bibitem{luo2024reasoninggraphsfaithfulinterpretable}
\BIBentryALTinterwordspacing
L.~Luo, Y.-F. Li, G.~Haffari, and S.~Pan, ``Reasoning on graphs: Faithful and interpretable large language model reasoning,'' 2024. [Online]. Available: \url{https://arxiv.org/abs/2310.01061}
\BIBentrySTDinterwordspacing

\bibitem{li2023graphreasoningquestionanswering}
\BIBentryALTinterwordspacing
S.~Li, Y.~Gao, H.~Jiang, Q.~Yin, Z.~Li, X.~Yan, C.~Zhang, and B.~Yin, ``Graph reasoning for question answering with triplet retrieval,'' 2023. [Online]. Available: \url{https://arxiv.org/abs/2305.18742}
\BIBentrySTDinterwordspacing

\bibitem{fatemi2023talklikegraphencoding}
\BIBentryALTinterwordspacing
B.~Fatemi, J.~Halcrow, and B.~Perozzi, ``Talk like a graph: Encoding graphs for large language models,'' 2023. [Online]. Available: \url{https://arxiv.org/abs/2310.04560}
\BIBentrySTDinterwordspacing

\bibitem{10387715}
S.~Pan, L.~Luo, Y.~Wang, C.~Chen, J.~Wang, and X.~Wu, ``Unifying large language models and knowledge graphs: A roadmap,'' \emph{IEEE Transactions on Knowledge and Data Engineering}, vol.~36, no.~7, pp. 3580--3599, 2024.

\bibitem{10697304}
B.~Jin, G.~Liu, C.~Han, M.~Jiang, H.~Ji, and J.~Han, ``Large language models on graphs: A comprehensive survey,'' \emph{IEEE Transactions on Knowledge and Data Engineering}, vol.~36, no.~12, pp. 8622--8642, 2024.

\bibitem{Hager2024}
\BIBentryALTinterwordspacing
P.~Hager, F.~Jungmann, K.~Bhagat, I.~Hubrecht, M.~Knauer, J.~Vielhauer, R.~Holland, R.~Braren, M.~Makowski, G.~Kaisis, and D.~Rueckert, ``Evaluating and mitigating limitations of large language models in clinical decision making,'' \emph{medRxiv}, 2024. [Online]. Available: \url{https://www.medrxiv.org/content/early/2024/01/26/2024.01.26.24301810}
\BIBentrySTDinterwordspacing

\bibitem{brody2022how}
S.~Brody, U.~Alon, and E.~Yahav, ``How attentive are graph attention networks?'' in \emph{International Conference on Learning Representations}, 2022.

\bibitem{uwcmg_mygene2}
{University of Washington Center for Mendelian Genomics}, ``Mygene2,'' Retrieved from \url{https://www.mygene2.org}, n.d.

\bibitem{chandak2023building}
P.~Chandak, K.~Huang, and M.~Zitnik, ``Building a knowledge graph to enable precision medicine,'' \emph{Scientific Data}, vol.~10, no.~1, p.~67, 2023.

\bibitem{paszke2017automatic}
A.~Paszke, S.~Gross, S.~Chintala, G.~Chanan, E.~Yang, Z.~DeVito, Z.~Lin, A.~Desmaison, L.~Antiga, and A.~Lerer, ``Automatic differentiation in pytorch,'' in \emph{NIPS-W}, 2017.

\bibitem{fey2019fast}
M.~Fey and J.~E. Lenssen, ``Fast graph representation learning with {PyTorch Geometric},'' in \emph{ICLR Workshop on Representation Learning on Graphs and Manifolds}, 2019.

\end{thebibliography}

\newpage
\appendix
\section*{Limitations of Public Datasets and the Need for Robust Evaluation} \label{subsec:imbalance} 




\begin{table}[h]
\centering
\caption{Performance comparison of PhenoApt and Amelie on frequent genes in the MyGene2 dataset. All values are shown in percentages.}
\setlength{\tabcolsep}{1mm}
\small
\begin{tabular}{llrrrrr}
\toprule
\multirow{2}{*}{\textbf{Gene}} & \multirow{2}{*}{\textbf{Method}} & \multicolumn{4}{c}{\textbf{Performance}} & \multirow{2}{*}{\textbf{\#Patients}} \\
\cmidrule(lr){3-6}
& & Hits@1 & Hits@5 & Hits@10 & MRR & \\
\midrule
\multirow{2}{*}{PIEZO2}  
    & PhenoApt & 86.1 & 94.4 & 97.2 & 89.6 & 36 \\
    & Amelie   & 16.7 & 33.3 & 58.3 & 27.9 & 36 \\
\midrule
\multirow{2}{*}{NALCN}  
    & PhenoApt & 86.7 & 93.3 & 93.3 & 90.4 & 15 \\
    & Amelie   & 93.3 & 100.0 & 100.0 & 96.7 & 15 \\
\midrule
\multirow{2}{*}{SF3B4}  
    & PhenoApt & 40.0 & 86.7 & 93.3 & 57.8 & 15 \\
    & Amelie   & 53.3 & 66.7 & 80.0 & 61.3 & 15 \\
\midrule
\multirow{2}{*}{MYH3}   
    & PhenoApt & 20.0 & 70.0 & 80.0 & 44.4 & 10 \\
    & Amelie   & 20.0 & 30.0 & 60.0 & 30.5 & 10 \\
\bottomrule
\end{tabular}
\label{tab:per_gene_comparison_full}
\end{table}

While MyGene2 remains a widely used benchmark in rare disease gene prioritization, its structural limitations-such as the dominance of a few causal genes-create biases that inflate model performance. This is particularly problematic for methods relying on frequency patterns or prior knowledge tied to common genes.

\noindent To address these limitations, recent simulated datasets (e.g., based on UDN patients~\cite{10.1093/nar/gkaa1043}) have been proposed as privacy-preserving alternatives. These datasets include more diverse causal genes and phenotype profiles, introducing realistic noise and uncertainty that better reflect clinical use cases.

\noindent A key distinction of the simulated test set is its more balanced representation of genes from the knowledge graph. For example, common MyGene2 genes are largely absent or underrepresented: PIEZO2 is entirely missing, NALCN appears in only 4 cases, and SF3B4 in 6. Consequently, methods like Amelie and PhenoApt-whose performance on MyGene2 is largely driven by these genes-exhibit substantial drops in accuracy: for instance, NALCN MRR drops to 0.6\% (PhenoApt) and 9.9\% (Amelie); for SF3B4, to 7.7\% and 22.5\%, respectively.

\noindent This degradation is not solely due to missing frequent genes. The simulated cohort presents a greater variety of phenotypes per patient, many of which are noisy or non-specific, mimicking the challenges of real-world diagnostic data extracted from unstructured clinical notes. Frequency-based shortcuts become ineffective, and models are required to generalize under higher uncertainty.

\noindent Overall, the simulated evaluation setup acts as a more stringent testbed for assessing robustness and generalizability, especially for prioritizing rare or under-characterized genes that are clinically most valuable.

\section{Visualization of Subgraph-Based Successes and Failures}

\begin{figure}[h!]
    \centering
    \includegraphics[width=0.95\columnwidth]{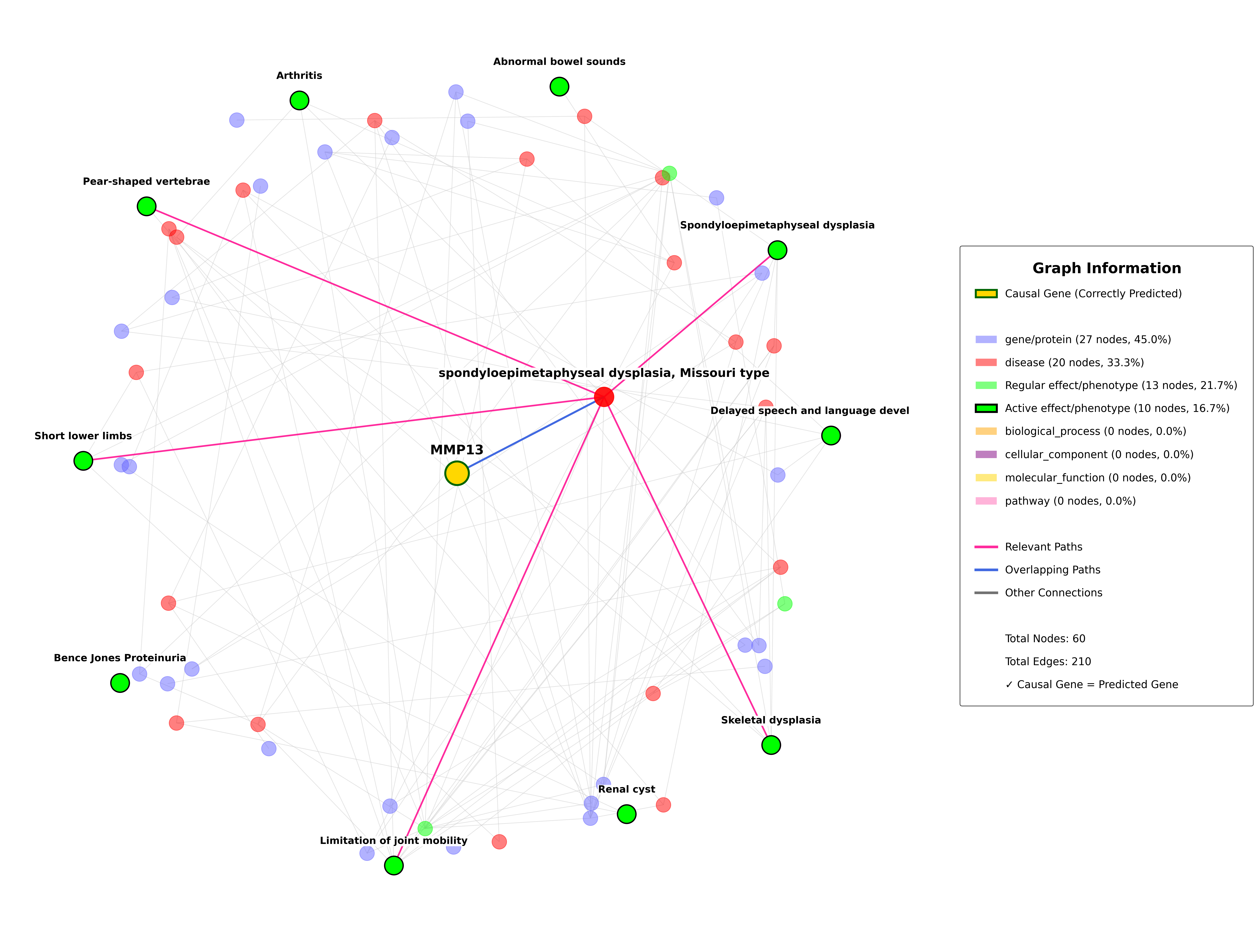}
    \caption{Patient 8 with 10 active phenotypes. Subgraph contains 60 nodes and 210 edges. Causal gene correctly predicted.}
    \label{fig:patient_subgraphA}
\end{figure}

\noindent\Cref{fig:patient_subgraphA} shows the subgraph for a patient with ten active phenotypes, where RareNet correctly predicts the causal gene. The model identifies a cohesive structure in which the majority of phenotype nodes are connected through a shared disease node before reaching the causal gene. The subgraph retains around $60$ nodes and $210$ edges from the full knowledge graph. Although some phenotype nodes remain more peripheral, they are included for completeness, reflecting the ability of RareNet to balance focus and coverage in its subgraph extraction.

\begin{figure}[h]
    \centering
    \includegraphics[width=0.95\columnwidth]{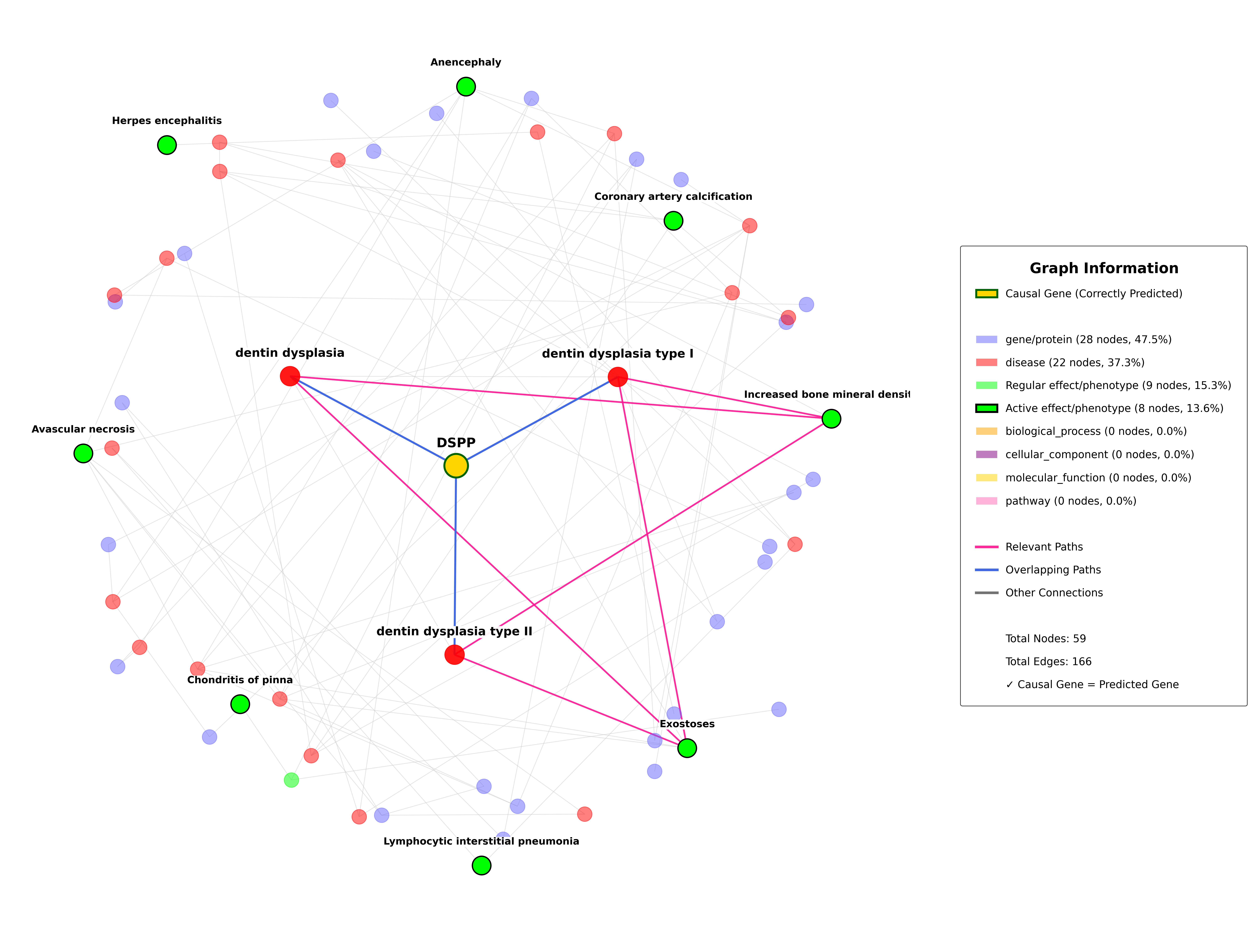}
    \caption{Patient 565 with 8 active phenotypes. Subgraph contains 50 nodes and 175 edges. Causal gene correctly predicted.}
    \label{fig:patient_subgraphB}
\end{figure}

\noindent Figure~\ref{fig:patient_subgraphB} shows a second correct-prediction example, with eight active phenotypes and a more compact subgraph structure. The predicted causal gene is reached via a tighter cluster of disease nodes (red). Several phenotypes converge through shared edges (blue), while others form more isolated routes. The overall subgraph contains approximately $50$ nodes, showing that RareNet can adapt to patient complexity while maintaining interpretability. Even phenotypes that are not directly connected to the gene are retained to provide a full picture of the context of the patient. These next two figures illustrate scenarios in which the top prediction of RareNet was incorrect, yet the causal gene remained part of the subgraph. In real diagnostic workflows, this can still aid a clinician in verifying or refuting the proposed gene of the model.

\begin{figure}[h]
    \centering
    \includegraphics[width=0.95\columnwidth]{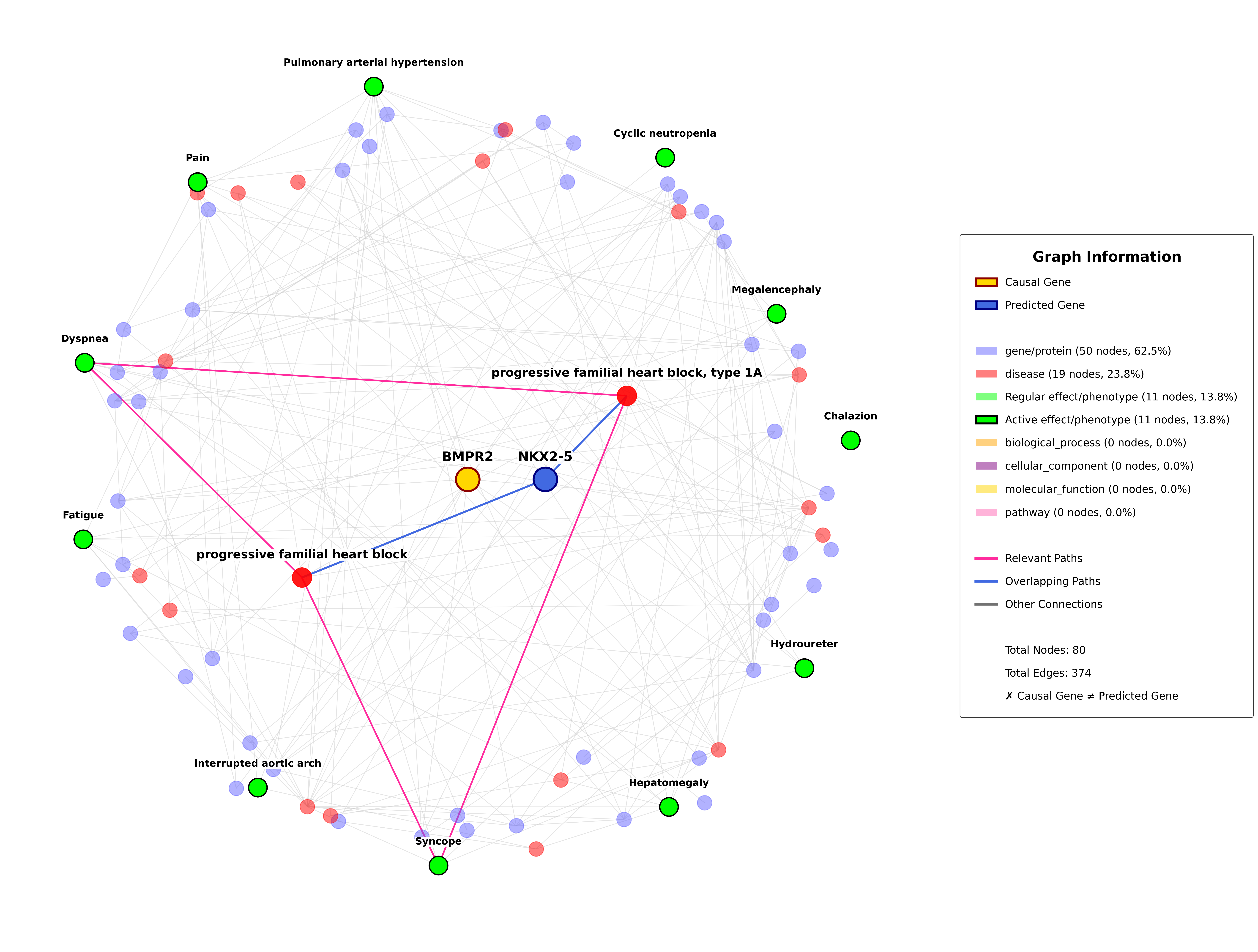}
    \caption{Patient 148 with 10 active phenotypes. Subgraph contains 54 nodes and 198 edges. Causal gene not correctly predicted (predicted gene shown in blue with thick blue border). Connections to the predicted gene are sparse, with fewer strong or overlapping paths from the phenotypes.}
    \label{fig:patient_subgraphC}
\end{figure}

\noindent Figure~\ref{fig:patient_subgraphC} depicts a patient with ten active phenotypes for which the predicted top gene (blue with black thick borders) is not the causal gene. Phenotypes still link to the candidate gene of the model by pink or blue edges, but the connections are less extensive than in the correct cases. This allows a clinician or researcher to spot potential weaknesses in the route to the predicted gene, there may be fewer overlapping or strongly weighted edges supporting that choice.

\noindent Figure~\ref{fig:patient_subgraphC} depicts a patient with ten active phenotypes for which the predicted top gene (blue with thick black borders) is not the causal gene. Phenotypes still link to the candidate gene of the model by pink or blue edges, but the connections are less extensive than in the correct cases. In this case, the incorrectly predicted gene is functionally or semantically close to the causal gene (e.g., involved in a related biological pathway), highlighting that some mispredictions by RareNet may still provide clinically relevant suggestions. Importantly, even when the top prediction is incorrect, the extracted subgraph can remain useful: it enables clinicians or researchers to analyze alternative candidates, as the causal gene is still included in the subgraph.

\begin{figure}[h]
    \centering
    \includegraphics[width=0.95\columnwidth]{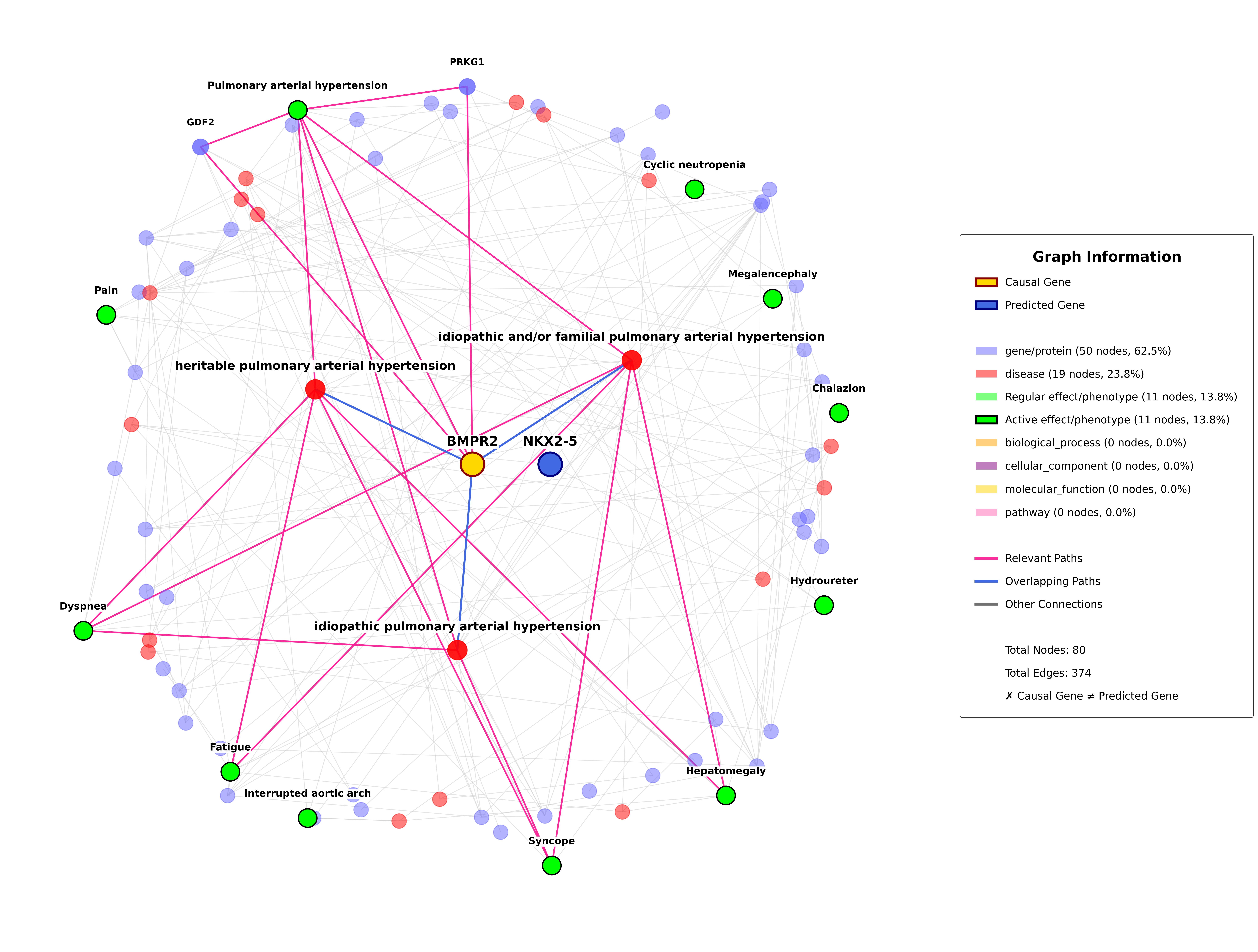}
    \caption{Patient 148 with 10 active phenotypes. Same subgraph as previous, highlighting the causal gene (yellow). Causal gene not predicted as top-1, but included in subgraph. Stronger and more overlapping edges (blue) from the phenotypes suggest a more plausible explanation.}
    \label{fig:patient_subgraphD}
\end{figure}

\noindent Figure~\ref{fig:patient_subgraphD} shows the causal gene for the same patient case. Although RareNet did not rank it as the top prediction, this correct gene was still included in the subgraph. Upon selecting it in a visualization tool, the clinician or researcher would see denser and more strongly weighted paths (indicated by thicker or overlapping edges) from the active phenotypes. This demonstrates that the subgraph provides meaningful context: even when the model's top prediction is weakly supported, users can leverage the richer connectivity of the causal gene to identify more plausible alternatives. In this example, the predicted gene is not functionally similar to the causal gene, despite phenotypic overlap, indicating that the model can sometimes fail to capture deeper biological relationships. These visualizations highlight both the strengths and the interpretability limits of the approach: while RareNet often retrieves biologically meaningful candidates and enables further analysis through its extracted subgraphs, it may occasionally suggest genes that, although phenotypically relevant, are functionally unrelated to the causal cause. This underlines the importance of expert validation in practical use.

\end{document}